\documentclass[runningheads]{llncs}
\usepackage{graphicx}
\usepackage{amsmath}
\usepackage{multirow}
\usepackage{array}
\usepackage{multicol}
\DeclareMathOperator{\Tr}{Tr}
\DeclareMathOperator*{\argmin}{arg\,min}
\usepackage{array, boldline, makecell, booktabs}

\usepackage[ruled,vlined]{algorithm2e}
\begin{document}
\title{
A Deep-Generative Hybrid Model to Integrate Multimodal and Dynamic Connectivity for Predicting Spectrum-Level Deficits in~Autism}
% % abbreviated author list (for running head)
\authorrunning{Niharika Shimona D'Souza et al.} % abbreviated author list (for running head)
\author{Niharika Shimona D'Souza\inst{1} 
 \textsuperscript{*}
\and Mary Beth Nebel~\inst{2}~\inst{3} \and Deana Crocetti~\inst{2} \and Nicholas Wymbs~\inst{2}\inst{3} \and Joshua Robinson \inst{2} \and Stewart Mostofsky\inst{2}\inst{3}\inst{4} \and Archana~Venkataraman\inst{1} }

 % index{D'Souza,Niharika S.}
 % index{Nebel,Mary Beth}
 % index{Crocetti,Deana}
 % index{Wymbs,Nicholas}
 % index{Robinson,Joshua}
 % index{Mostofsky,Stewart}
 % index{Venkataraman,Archana}

\institute{Dept. of Electrical and Computer Eng., Johns Hopkins University, Baltimore, USA
\email{\textsuperscript{*}}{Shimona.Niharika.Dsouza@jhu.edu}
\and
Center for Neurodevelopmental \& Imaging Research, Kennedy Krieger Institute
\and
Dept. of Neurology, Johns Hopkins School of Medicine, Baltimore, USA
\and
Dept. of Psychiatry \& Behavioral Science, Johns Hopkins School of Medicine, USA}

\maketitle              % typeset the header of the contribution
\begin{abstract}
We propose an integrated deep-generative framework, that jointly models complementary information from resting-state functional MRI (rs-fMRI) connectivity and diffusion tensor imaging (DTI) tractography to extract predictive biomarkers of a disease. The generative part of our framework is a structurally-regularized Dynamic Dictionary Learning (sr-DDL) model that decomposes the dynamic rs-fMRI correlation matrices into a collection of shared basis networks and time varying patient-specific loadings. This matrix factorization is guided by the DTI tractography matrices to learn anatomically informed connectivity profiles. The deep part of our framework is an LSTM-ANN block, which models the temporal evolution of the patient sr-DDL loadings to predict multidimensional clinical severity. Our coupled optimization procedure collectively estimates the basis networks, the patient-specific dynamic loadings, and the neural network weights. We validate our framework on a multi-score prediction task in $57$ patients diagnosed with Autism Spectrum Disorder (ASD). Our hybrid model outperforms state-of-the-art baselines in a five-fold cross validated setting and extracts interpretable multimodal neural signatures of brain dysfunction in ASD.
\end{abstract}
\section{Introduction}
\par Autism Spectrum Disorder (ASD) is a complex neurodevelopmental disorder characterized by impaired social communicative skills and awareness, coupled with restricted/repetitive behaviors. These symptoms and levels of disability vary widely across the ASD spectrum. Neuroimaging techniques such as rs-fMRI and DTI are gaining popularity for studying aberrant brain connectivity in ASD \cite{bennett2013advances}. Rs-fMRI allows us to assess the functional organization of the brain by tracking changes in steady-state co-activation \cite{lee2013resting}, while DTI measures structural connectivity via the diffusion of water molecules in the brain \cite{assaf2008diffusion}. However, the high data dimensionality, coupled with noise and patient variability, have limited our ability to integrate these modalities to understand behavioral deficits.
\par Techniques integrating structural and functional connectivity focus heavily on groupwise discrimination from the static connectomes. Methods include statistical tests on the node or edge biomarkers  \cite{skudlarski2008measuring}, data-driven representations \cite{sui2013combination,nandakumar2018defining}, and neural networks \cite{aghdam2018combination} for classification. However, none of these methods tackle continuous-valued prediction, e.g., quantifying level of deficit. Deep learning is becoming increasingly popular for continuous prediction. The work of \cite{kawahara2017brainnetcnn} proposes a specialized end-to-end convolutional network that predicts clinical outcomes from DTI connectomes. The authors of \cite{d2019integrating} combine a dictionary learning on the rs-fMRI correlations with an ANN to predict clinical severity in ASD patients. However, these methods focus on a single neuroimaging modality and do not leverage complementary information between structure and function. 
\par There is now growing evidence that functional connectivity between regions is a dynamically evolving process \cite{cabral2017functional}, and that modeling this evolution is crucial to understanding disorders like ASD \cite{price2014multiple,rashid2014dynamic}. Hence, recent methods have been proposed that use either a sparse decomposition of the rs-fMRI connectomes \cite{cai2017estimation}, or a temporal clustering for ASD/control discrimination \cite{rabany2019dynamic}. While promising, these approaches focus exclusively on rs-fMRI and ignore structural information.
\par We propose a hybrid deep-generative model that integrates structural and dynamic functional connectivity with behavior into a unified optimization framework. Our generative component is a structurally-regularized Dynamic Dictionary Learning (sr-DDL) model, which uses anatomical priors from DTI to regularize a time-varying decomposition of the rs-fMRI correlation matrices. Here, the connectivity profiles are explained by shared basis networks and time-varying patient-specific loadings. Simultaneously, these loadings are input to a deep network which uses an LSTM (Long Short Term Memory Network) to model temporal trends and an ANN (Artificial Neural Network) to predict clinical severity. Our optimization procedure learns the bases, loadings, and neural network weights most predictive of behavioral deficits in ASD. We obtain a representation which is both interpretable and generalizes to unseen patients, thus providing a comprehensive characterization of the disorder.
\section{A Deep-Generative Hybrid Model for Connectomics}
\begin{figure}[t!]
   \centering
   \includegraphics[width=\dimexpr \textwidth-6\fboxsep-6\fboxrule\relax]{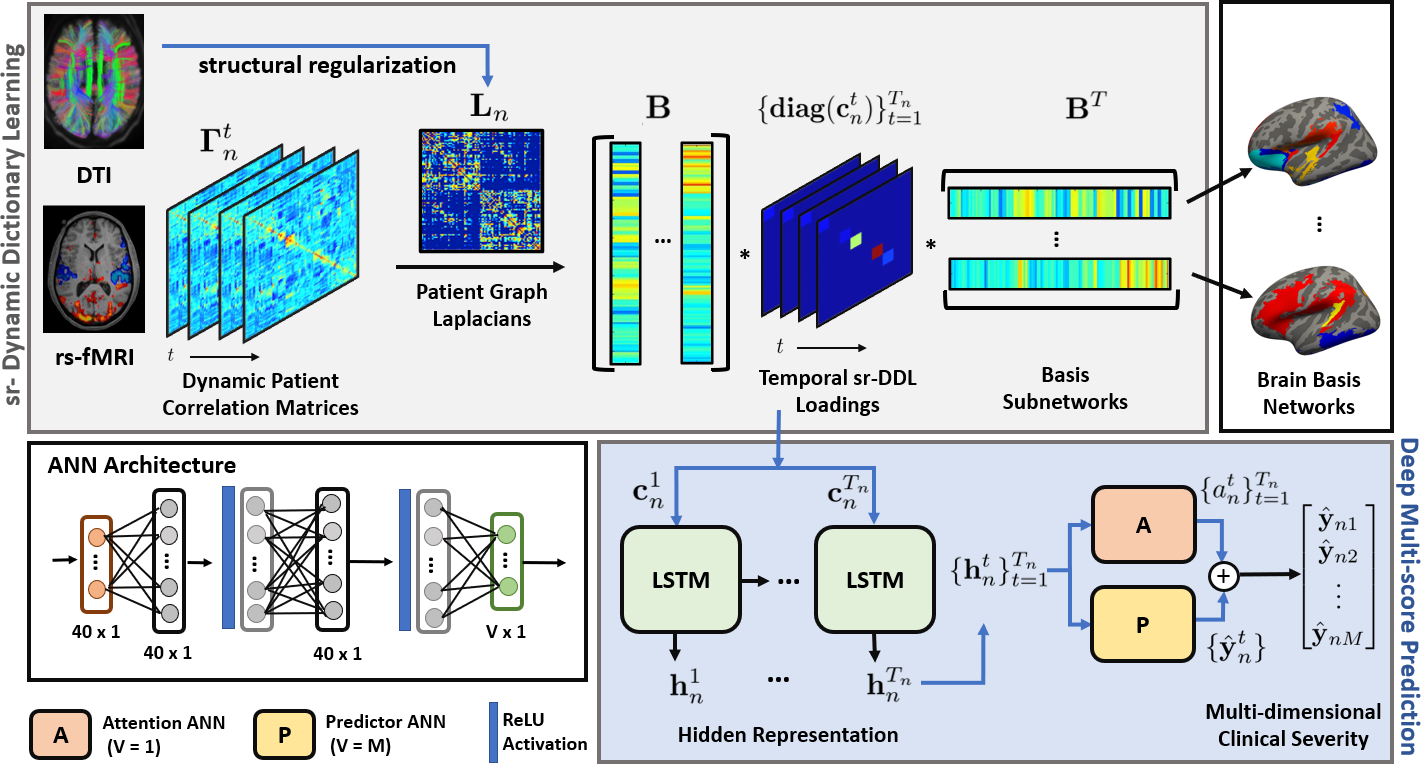}
   \caption{Framework to integrate structural \& dynamic functional connectivity for multi-task prediction \textbf{Gray Box:}~sr-DDL module for rs-fMRI dynamic correlation matrices and DTI connectivity matrices. \textbf{Blue Box:}~LSTM-ANN for multi-score prediction. }  \label{CCS}
\end{figure}
Fig.~\ref{CCS} illustrates the generative (sr-DDL) and deep (LSTM-ANN) components of our framework. Let $P$ be the number of ROIs in our brain parcellation and $N$ be the number of patients. The rs-fMRI dynamic correlation matrices for patient $n$ are denoted by $\{\mathbf{\Gamma}^{t}_{n}\}_{t=1}^{T_{n}}\in \mathcal{R}^{P \times P}$, with  $T_{n}$ being the number of time steps. $\mathbf{L}_{n} \in \mathcal{R}^{P \times P}$ is the corresponding DTI connectivity information, and $\mathbf{y}_{n} \in \mathcal{R}^{M \times 1}$ is a vector of $M$ concatenated severity measures. Given the inputs $\mathcal{P}=\{\{\mathbf{\Gamma}^{t}_{n}\},\mathbf{L}_{n},\mathbf{y}_{n}\}^{N}_{n=1}$, our framework optimizes the following joint objective:\begin{equation}
\mathcal{J}(\mathbf{B},\{\mathbf{c}^{t}_{n}\},\mathbf{\Theta};\mathcal{P}) = \underbrace{\sum_{n} \mathcal{D}(\mathbf{B},\{\mathbf{c}^{t}_{n}\};\{\mathbf{\Gamma}^{t}_{n}\},\mathbf{L}_{n})}_{\textbf{sr-DDL loss}} + \lambda \underbrace{\sum_{n}\mathcal{L}(\mathbf{\Theta},\{\mathbf{c}^{t}_{n}\};\mathbf{y}_{n})}_{\textbf{deep network loss}}
\label{JointObj}
\end{equation}
\noindent{\underline{\textbf{sr-DDL Factorization:}}} We represent the correlation matrices $\mathbf{\Gamma}^{t}_{n}$ by a shared basis $\mathbf{B} \in \mathcal{R}^{P\times K}$  that captures template patterns of co-activity and temporal loadings $\mathbf{c}^{t}_{n} \in \mathcal{R}^{K \times 1}$ that indicate their time-varying strength:
\begin{equation}
    \mathcal{D}(\mathbf{B},\{\mathbf{c}^{t}_{n}\};\{\mathbf{\Gamma}^{t}_{n}\},\mathbf{L}_{n}) = \sum_{t}{\frac{1}{T_{n}}}{\vert\vert{\mathbf{\Gamma}^{t}_{n} - \mathbf{B}\mathbf{diag}(\mathbf{c}^{t}_{n})\mathbf{B}^{T}}\vert\vert}_{\mathbf{L}_{n}} \ \ s.t. \ \mathbf{B}^{T}\mathbf{B} = \mathcal{I}_{K}
    \label{sr-DDL}
\end{equation}
Here, $K$ is the size of our basis, and $\mathbf{diag}(\mathbf{c}^{t}_{n})$ is a diagonal matrix based on the elements of $\mathbf{c}^{t}_{n}$, and  $\mathcal{I}_{K}$ is the identity matrix of size $K$. The positive semi-definiteness of $\{\mathbf{\Gamma}^{t}_{n}\}$ further implies that $\mathbf{c}^{t}_{n}$ is non-negative. The orthonormality constraint on $\mathbf{B}$  helps us learn uncorrelated sub-networks that explain the rs-fMRI data well and implicitly regularize the optimization. 
\par Notice that Eq.~(\ref{sr-DDL}) uses a weighted Frobenius norm, rather than the standard $\ell_{2}$ penalty. Mathematically, this norm is computed as $\vert\vert{\mathbf{X}}\vert\vert_{\mathbf{L}_{n}}= \Tr({\mathbf{X}^{T}\mathbf{L}_{n}\mathbf{X}})$ \cite{manton2003geometry,schnabel1983forcing}, with $\mathbf{X}= \mathbf{\Gamma}^{t}_{n} - \mathbf{B}\mathbf{diag}(\mathbf{c}^{t}_{n})\mathbf{B}^{T}$. The matrix $\mathbf{L}_{n} \in \mathcal{R}^{P\times P}$ in our case is the normalized graph Laplacian \cite{banerjee2008spectrum} derived from the DTI adjacency matrix for patient $n$. The DTI adjacency is $1$ if there is at least one tract between the corresponding regions, and $0$ otherwise. Essentially, this structural-regularization encourages the functional decomposition to focus on explaining the functional connectivity between regions with an a-priori anatomical connection.

\noindent\underline{\textbf{{Deep Network:}}} The patient coefficients $\mathbf{c}^{t}_{n}$ are input to an LSTM-ANN network to predict the scores $\mathbf{y}_{n}$. The LSTM generates a hidden representation $\mathbf{h}^{t}_{n}$ over time. From here, the Predictor ANN (P-ANN) outputs a time varying estimate of the scores $\{\hat{\mathbf{y}}^{t}_{n}\}^{T_{n}}_{t=1}$. The Attention ANN (A-ANN) generates $T_{n}$ scalars, which we softmax across time to obtain the attention weights: $\{a_{n}^{t}\}^{T_{n}}_{t=1}$. These weights determine which time points for each patient are most relevant for behavioral prediction. The final prediction is an attention-weighted average across the estimates $\hat{\mathbf{y}}^{t}_{n}$. We~use~an~MSE~loss~in~Eq.~(\ref{JointObj})~to~obtain:
\begin{equation}
\mathcal{L}(\{\mathbf{c}^{t}_{n}\},\mathbf{y}_{n};\mathbf{\Theta})  =  {\vert\vert{\mathbf{\hat{y}}_{n}-\mathbf{{y}}_{n}}\vert\vert}^{2}_{F} = {\Bigg|\Bigg|{{\sum^{T_n}_{t}{\mathbf{\hat{y}}^{t}_{n} a^{t}_{n}}-\mathbf{{y}}_{n}}\Bigg|\Bigg|}^{2}_{F}}
\label{MSE}        
\end{equation}
\par We employ a two layered LSTM with hidden layer width $40$. As seen in Fig~\ref{CCS}, both the P-ANN and the A-ANN have two hidden layers with width $40$ with ReLU activations, with output size $(V)$ as $M$ and $1$ respectively.  We observed that these modeling choices are robust to issues with saturation and vanishing gradients that can hinder the training of deep neural networks. Finally, we fix the trade-off between the losses in Eq.~(\ref{JointObj}) at $\lambda=3$, and the number of networks to $K=15$ based on a grid search.
\subsection{Coupled Optimization Strategy:}
We use alternating minimization to optimize Eq.~(\ref{JointObj}) with respect to $\{\mathbf{B},\{\mathbf{c}^{t}_{n}\},\mathbf{\Theta}\}$. Here, we iteratively cycle through the updates for the dictionary $\mathbf{B}$, loadings $\{\mathbf{c}^{t}_{n}\}$, and the LSTM-ANN weights $\mathbf{\Theta}$ to obtain a \textit{joint solution}.
\par We note that there is a closed-form Procrustes solution for quadratic objectives \cite{everson1998orthogonal}. However, Eq.~(\ref{JointObj}) is bi-quadratic in $\mathbf{B}$, so it cannot be directly applied. Therefore, we adopt the strategy in \cite{d2018generative,d2019coupled,d2019integrating,d2020joint}, by which we introduce the constraints of the form $\mathbf{D}^{t}_{n} = \mathbf{B}\mathbf{diag}(\mathbf{c}^{t}_{n})$, with corresponding augmented Lagrangian variables $\{\mathbf{\Lambda}^{t}_{n}\}$. Thus, our objective from Eq.~(\ref{JointObj}) now becomes:
\begin{multline}
\mathcal{J}_{c} = \sum_{n,t}{\frac{1}{T_{n}}}{\vert\vert{\mathbf{\Gamma}^{t}_{n} - \mathbf{D}^{t}_{n}\mathbf{B}^{T}}\vert\vert}_{\mathbf{L}_{n}} + \lambda \sum_{n}\mathcal{L}(\mathbf{\Theta},\{\mathbf{c}^{t}_{n}\};\mathbf{y}_{n}) \ \  s.t.  \ \ \mathbf{B}^{T}\mathbf{B} = \mathcal{I}_{K} \\ 
+ \sum_{n,t}{\frac{\gamma}{T_{n}}\Big[{\Tr{\left[{(\mathbf{\Lambda}^{t}_{n})^{T}({\mathbf{D}^{t}_{n}-\mathbf{B}\mathbf{diag}(\mathbf{c}^{t}_{n})})}\right]}}+{{\frac{1}{2}}~{\vert\vert{\mathbf{D}^{t}_{n}-\mathbf{B}\mathbf{diag}(\mathbf{c}^{t}_{n})}\vert\vert}_{F}^{2}}}\Big]    
\label{const}
\end{multline}
\begin{figure}[t!]
   \centering
   \includegraphics[width=\dimexpr \textwidth-10\fboxsep-10\fboxrule\relax]{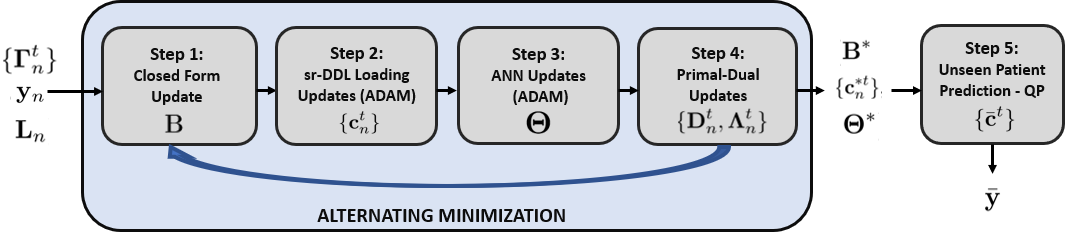}
   \caption{{Alternating minimization strategy for joint optimization of Eq.~(\ref{const}})  \label{AltMin}}
\end{figure}
Fig.~{\ref{AltMin}} outlines our coupled optimization strategy, with  steps detailed as follows:   

\medskip\noindent{\textbf{{Step~{1}: Closed form solution for $\mathbf{B}$.}}} Notice that Eq.~(\ref{const}) reduces to a Procrustes objective $\mathbf{B}^{*} = \argmin_{\mathbf{B} : \ \mathbf{B}^{T}\mathbf{B}=\mathcal{I}_{K}}{{\vert\vert{\mathbf{M}-\mathbf{B}}\vert\vert}^{2}_{F}}$ where: \begin{equation*}
    \mathbf{M} = \sum_{n}{\frac{1}{T_{n}}}\sum_{t}{(\mathbf{\Gamma}^{t}_{n}\mathbf{L}_{n}+ \mathbf{L}_{n}\mathbf{\Gamma}^{t}_{n})\mathbf{D}^{t}_{n}+ \gamma\mathbf{D}^{t}_{n}\mathbf{diag}(\mathbf{c}^{t}_{n})+ \gamma\mathbf{\Lambda}^{t}_{n}\mathbf{diag}(\mathbf{c}^{t}_{n})} 
\end{equation*} Given the singular value decomposition $\mathbf{M} = \mathbf{U}\mathbf{S}\mathbf{V}^{T}$, then $\mathbf{B}^{*} = \mathbf{U}\mathbf{V}^{T}$. Thus, $\mathbf{B}$ spans the anatomically weighted space of patient correlation matrices. 

\medskip\noindent\textbf{{Step~{2}: Updating the sr-DDL loadings  $\{\mathbf{c}_{n}^{t}\}$.}} The objective $\mathcal{J}_{c}$ in Eq.~(\ref{const}) decouples across patients. We can also incorporate the non-negativity constraint $\mathbf{c}^{t}_{nk} \geq 0$ by passing an intermediate vector $\hat{\mathbf{c}}^{t}_{n}$ through a ReLU. The ReLU pre-filtering allows us to optimize an unconstrained version of Eq.~(\ref{const}), which can be done via the stochastic ADAM algorithm \cite{kingma2015adam}. In essence, this optimization couples the parametric gradient from the augmented Lagrangians with the backpropagated gradient from the deep network (defined by fixed $\mathbf{\Theta}$). After convergence, the thresholded loadings ${\mathbf{c}}^{t}_{n} = ReLU(\hat{\mathbf{c}}^{t}_{n})$ are used in subsequent~steps.  

\medskip\noindent\textbf{{Step~{3}: Updating the Deep Network weights $\mathbf{\Theta}$}.}  We use backpropagation on the loss $\mathcal{L(\cdot)}$ to solve for $\mathbf{\Theta}$. Notice that we can handle missing clinical data by dropping the contributions of the unknown value of $\mathbf{y}_{nm}$ to the network during backpropagation. We use the ADAM \cite{kingma2015adam} optimizer with random initialization, a learning rate of $10^{-4}$, scaled by $0.95$ every $5$ epochs, and batch-size $1$.

\medskip\noindent\textbf{{Step~{4}: Updating the Constraint Variables $\{\mathbf{D}^{t}_{n},\mathbf{\Lambda}^{t}_{n}\}$.}}
We perform parallel primal-dual updates for the constraint pairs $\{\mathbf{D}^{t}_{n},\mathbf{\Lambda}^{t}_{n}\}$ \cite{afonso2010augmented}. Here, we cycle through the closed form update for $\mathbf{D}_{n}^{t}$ and gradient ascent for $\mathbf{\Lambda}^{t}_{n}$ until convergence.

\medskip\noindent\textbf{{Step~{5}: Prediction on Unseen Data.}}
In our cross-validated setting, we need to compute the sr-DDL loadings $\{\mathbf{\Bar{c}}^{t}\}_{t=1}^{\bar{T}}$ for a new patient based on the training $\mathbf{B}^{*}$. Since we do not know the score $\mathbf{\bar{y}}$ for this patient, we remove the contribution $\mathcal{L}(\cdot)$ from Eq.~(\ref{const}) and assume the constraints $\bar{\mathbf{D}}^{t} = \mathbf{B^{*}}\mathbf{diag}(\bar{\mathbf{c}}^{t})$ hold with equality, thus removing the Lagrangian terms. Essentially, the optimization for $\{\mathbf{\Bar{c}}^{t}\}$ reduces to decoupled quadratic programming (QP)~objectives~$\mathcal{Q}_{t}$~across~time: \begin{multline}
\ \ \ \ \ \ \ \ \ \ \ \ \Bar{\mathbf{c}}^{*t} =  \argmin_{\bar{\mathbf{c}}^{t}}{\frac{1}{2}}~{(\mathbf{\Bar{c}}^{t})^{T}\mathbf{\Bar{H}}\mathbf{\Bar{c}}^{t}} + \mathbf{\Bar{f}}^{T}\mathbf{\Bar{c}}^{t} \ \ s.t. \ \ \mathbf{\Bar{A}}\mathbf{\Bar{c}}^{t} \leq \mathbf{\Bar{b}} \ \ \ \ \ \ \ \ \ \ \ \ \ \   \\ 
\mathbf{\Bar{H}} = 2(\mathbf{B^{*}}^{T}\bar{\mathbf{L}}\mathbf{B^{*}}); \ \ \  \mathbf{\Bar{f}} = -[(\mathbf{B^{*}}^{T}(\bar{\mathbf{\Gamma}}\bar{\mathbf{L}}+ \bar{\mathbf{L}}\bar{\mathbf{\Gamma}})\mathbf{B^{*}})\circ \mathcal{I}_{K}]\mathbf{1}; \ \  \mathbf{\Bar{A}} = -\mathcal{I}_{K}  \notag 
 \   \mathbf{\Bar{b}} = \mathbf{0} 
\end{multline}
Where, $\circ$ denotes the Hadamard product. Finally, we  estimate $\bar{\mathbf{y}}$ via a forward pass through the LSTM-ANN. 
\subsection{Baseline Comparisons:}
We compare the predictive performance of our framework against three baselines:
\begin{itemize}
\item[1.]{Two channel BrainNet CNN \cite{kawahara2017brainnetcnn} on static rs-fMRI and DTI connectomes }
\item[2.]{PCA on DTI weighted dynamic rs-fMRI correlation features + LSTM-ANN }
\item[3.]{Decoupled sr-DDL factorization followed by the LSTM-ANN}
\end{itemize}
\par The first baseline integrates multi-modal DTI connectivity with static rs-fMRI connectivity via the BrainNet CNN introduced in \cite{kawahara2017brainnetcnn}. The original architecture is designed to predict cognitive scores from DTI. Here, we modify the BrainNet CNN to have two branches, one for rs-fMRI patient correlation matrices ${\mathbf{\Gamma}}_{n} \in \mathcal{R}^{P \times P} $ and the other for the DTI Laplacians $\mathbf{L}_{n} \in \mathcal{R}^{P \times P}$. We also modify the ANN in \cite{kawahara2017brainnetcnn} to pool the learned representations and predict $M$ clinical severity measures. The hyperparameters are fixed according to \cite{kawahara2017brainnetcnn}. 
\par For the second baseline, we weight the dynamic ${P\times (P-1)}/2$ rs-fMRI correlation features by the respective DTI Laplacian features. We then use Principal Component Analysis (PCA) to reduce the data dimensionality to $K =15$, followed by a similar LSTM-ANN framework to map onto behavior. 
\par Finally, we examine the score prediction upon excluding the DTI regularization from our deep-generative hybrid. This helps us evaluate the advantage of our multi-modal data integration, as opposed to analyzing rs-fMRI data alone. 
\section{Experimental Evaluation and Results:}
\medskip\noindent\textbf{{Data and Preprocessing.}} We validate our framework on a cohort of $57$ children with high-functioning ASD. Rs-fMRI and DTI scans are acquired on a Philips $3T$ Achieva scanner (\textbf{rs-fMRI:} EPI, TR/TE = $2500/30$ms, flip angle = $70$, res = $3.05\times3.15\times3$mm, duration = $128$ or $156$ time samples; \textbf{DTI}: EPI, SENSE factor = $2.5$, TR/TE = $6356/75$ms, res = $0.8\times 0.8 \times 2.2$mm, b-value = $700$s/$mm^{2}$, $32$ gradient directions). Rs-fMRI data was preprocessed through a standard pipeline that included motion correction, normalization to the MNI template, spatial and temporal filtering, and nuisance regression with CompCorr \cite{behzadi2007component}. DTI data was preprocessed using the FDT pipeline in FSL \cite{jenkinson2012fsl}. We perform tractography using the BEDPOSTx and PROBTRACKx functions in FSL \cite{behrens2007probabilistic}. 
\par We use the Automatic Anatomical Labeling (AAL) atlas \cite{tzourio2002automated} to define $116$ brain ROIs. A sliding window protocol (length=$45$, stride=$5$) was used to extract dynamic rs-fMRI correlations matrices. We subtract the first eigenvector, which is a roughly constant bias, and use the residual matrices as the inputs $\{\mathbf{\Gamma}^{t}_{n}\}$ for all methods. The DTI connectivity matrix is binary, where $1$ corresponds to at least one tract between the two regions. We impute the DTI connectivity for the $11$ patients, who do not have DTI based on the training data.
\par We rely on three clinical measures to characterize various impairments associated with ASD. The Autism Diagnostic Observation Schedule (ADOS) \cite{payakachat2012autism} captures socio-communicative deficits and restricted/repetitive behaviors via clinician evaluation (dynamic range: $0-30$). The Social Responsiveness Scale (SRS) \cite{payakachat2012autism} quantifies impaired social functioning via a parent/teacher questionnaire (dynamic range: $70-200$). Finally, Praxis \cite{dziuk2007dyspraxia,mostofsky2006developmental} measures the ability to perform skilled motor gestures on command. A videotaped performance of the child is scored by two research-reliable raters (dynamic~range:~$0-100$).

\begin{figure}[t!]    % start subfigure 3
    \centering
      \includegraphics[width=\dimexpr \textwidth\fboxsep-6\fboxrule\relax]{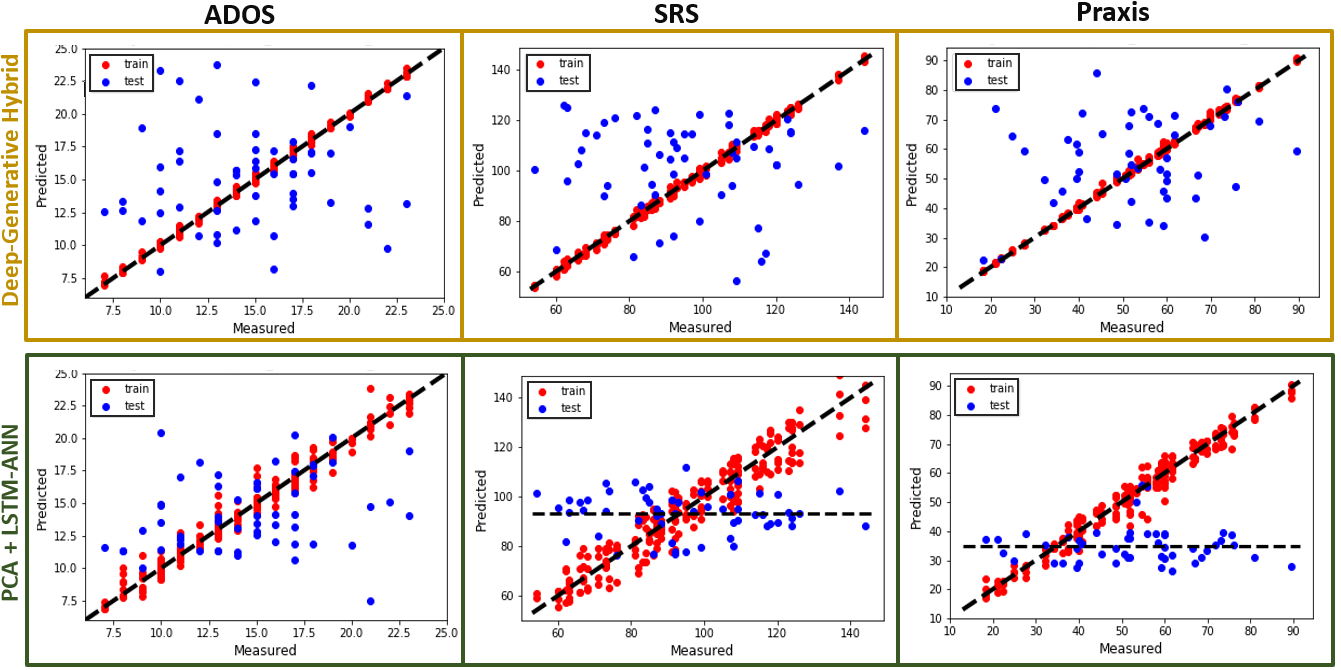}
   \caption{Multi-Score Prediction by \textbf{Left:} ADOS \textbf{Middle:} SRS \textbf{Right:} Praxis by \textbf{Yellow Box:} deep-generative hybrid. \textbf{Green Box:} PCA+LSTM-ANN}\label{Multi-Score}
\end{figure}

\medskip\noindent\textbf{{Multi-dimensional Severity Prediction.}}
1Fig.~\ref{Multi-Score} contrasts the performance of the deep-generative hybrid against the best performing baseline. We have included performance comparisons against the remaining baselines in Fig. 1 of the Supplementary Document. Here, we plot the severity score as given by the clinician on the $\mathbf{x}$-axis, and the score predicted by the algorithm on the $\mathbf{y}$-axis. The training and testing performance is illustrated by the red and blue points, respectively. The bold $\mathbf{x} = \mathbf{y} $ diagonal line indicates ideal performance. Notice that all methods have a good training fit for all the scores. However, in case of testing performance, our method outperforms the baselines in almost all cases. Empirically, we are able to tune the baseline hyperparameters to obtain good testing performance on a single score (e.g. ADOS for Baseline~$2$), but the prediction of the remaining scores suffer. In contrast, the testing predictions from our framework follow the diagonal line more closely for all the scores. We believe that the representational flexibility of our deep network along with the joint optimization helps us generalize well.
\begin{table}[t!]
\centering
{\caption{Performance based on \textbf{Median Absolute Error (MAE)} and \textbf{Mutual Information (MI)}. Lower MAE and higher MI indicate better performance.}\label{table:1}}
\begin{tabular}{|c |c | c| c| c| c|} 
\hline
  \textbf{Score} &\textbf{Method} &\textbf{MAE Train} & \textbf{MAE Test} & \textbf{MI Train} & \textbf{MI Test} \\  
\Xhline{2\arrayrulewidth}
  \multirow{4}{4em}{ADOS} & BrainNetCNN & 1.90 & 3.50 & 0.96 & 0.25\\
 & PCA \& LSTM-ANN & 0.34 & \textbf{2.47} & 0.96 & \textbf{0.35} \\
 & Without DTI reg. & 0.13 & 3.27 & 0.99 & 0.26\\
 & \textbf{Our Framework} & \textbf{0.08}& \underline{2.84}& \textbf{0.99}& \underline{0.34}\\
[0.2ex]  
\hline
 \multirow{4}{4em}{SRS} & BrainNetCNN & 5.25 & 18.96 & 0.83 & 0.75 \\
 & PCA \& LSTM-ANN & 4.73  & 19.05 & 0.95 & 0.68\\
 & Without DTI reg. & \textbf{0.49} & 18.70 & 0.97 & 0.77\\
 & \textbf{Our Framework} & \underline{0.51}& \textbf{17.81}& \textbf{0.98}& \textbf{0.88}
 \\ [0.2ex]
 \hline
 \multirow{4}{4em}{Praxis} & BrainNetCNN & 3.78 & 15.15 & 0.95 & 0.19 \\
 & PCA \& LSTM-ANN & 2.21 & 20.71 & 0.90 & 0.47\\
 & Without DTI reg. & 1.09 & 17.34 & 0.99 & 0.49\\
 & \textbf{Our Framework} & \textbf{0.13}& \textbf{13.50}& \textbf{0.99}& \textbf{0.85}\\
 [0.4ex]
 \hline
\end{tabular}
\end{table}

\begin{figure}[t!]
 \centerline{\includegraphics[width=\dimexpr \textwidth-6\fboxsep-10\fboxrule\relax]{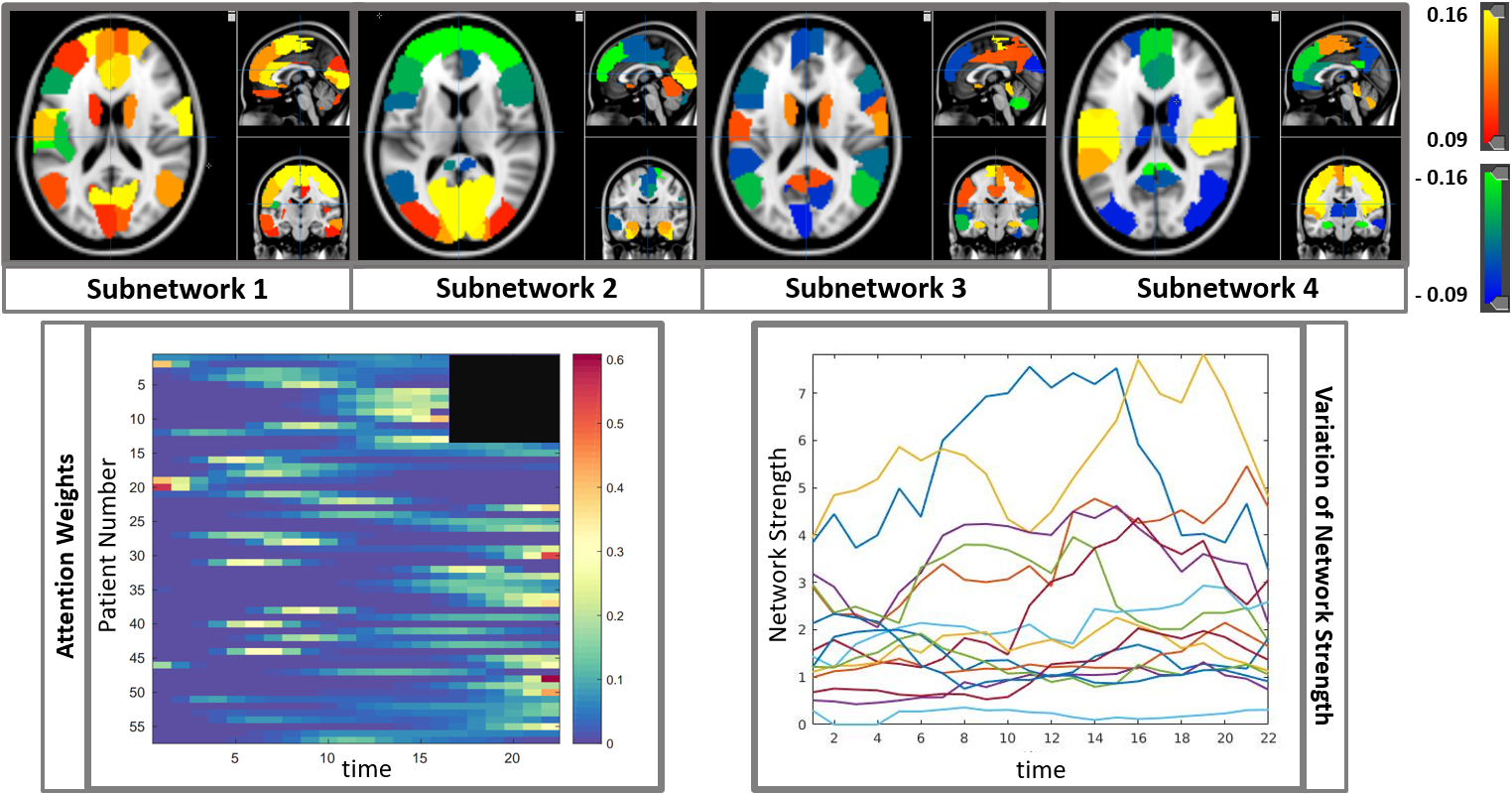}}
{\caption{ \textbf{Top:} Subnetworks identified by the deep-generative hybrid. The red and orange regions are anti-correlated with the blue and green regions
\textbf{Bottom:} \textbf{(Left)} Learned attention weights \textbf{(Right)} Variation of network strength over time }\label{ADOS_NL}} 
\end{figure}

\medskip\noindent\textbf{{Clinical Interpretability.}} Fig.~\ref{ADOS_NL} (top) illustrates four representative subnetworks learned in $\mathbf{B}$. (We have included the complete set of sub-network characterizations in Fig. 2 in the Supplementary Document.) Regions storing positive values are anticorrelated with negative regions. Subnetwork~$1$ includes regions from the default mode network (DMN), which has been widely reported in ASD \cite{nebel2016intrinsic}.  Subnetwork~$2$ exhibits contributions from higher order visual processing and sensorimotor areas, concurring with behavioral reports of reduced visual-motor integration in ASD \cite{nebel2016intrinsic}. Subnetworks~$3$ exhibits contributions from the central executive control network and insula, believed to be essential for switching between goal-directed and self-referential behavior \cite{sridharan2008critical}. Subnetwork~$4$ includes prefrontal and DMN regions, along with subcortical areas: associated with social-emotional regulation \cite{pouw2013link}.
\par Fig.~\ref{ADOS_NL} (bottom left) illustrates the learned attentions output by the A-ANN for all $57$ patients during testing. We group patients with shorter scans in the first few rows of the plot. We have blackened the rest of the time points for these patients. The colorbar indicates the strength of the attention weights. The flagged non-zero weights denote intervals of the scan considered especially relevant for prediction. We observe that the network highlights the start of the scan for several patients, while it prefers focusing on the end of the scan for some others. This is indicative of the underlying patient heterogeneity.
\par Lastly, we illustrate the variation of the network strength for a patient in the cohort over the scan duration in Fig.~\ref{ADOS_NL} (bottom right). Each solid colored line corresponds to one of the $15$ sub-networks. Over the scan duration, each network cycles through phases of activity and relative inactivity. Thus, only a few networks at each time step contribute to the patient's dynamic connectivity profile. This parallels the transient brain-states hypothesis in dynamic rs-fMRI connectivity \cite{damaraju2014dynamic}, with active states as corresponding sub-networks in $\mathbf{B}$.
\section{Conclusion}
We have introduced a novel deep-generative framework to integrate complementary information from the functional and structural neuroimaging domains, and simultaneously explain behavioral deficits in ASD. Our unique structural regularization elegantly injects anatomical information into the rs-fMRI functional decomposition, thus providing us with an interpretable brain basis. Our LSTM-ANN term not only models the temporal variation, but also helps isolate key dynamic resting-state signatures, indicative of clinical impairments. Our coupled optimization procedure ensures that we learn effectively from limited training data and generalize well to unseen patients. Finally, our framework makes very few assumptions and can potentially be applied to study other neuro-psychiatric disorders (eg. ADHD, Schizophrenia) as an effective diagnostic tool.

\bibliographystyle{splncs04}

{{\bibliography{MyRefs.bib}}}

\begin{thebibliography}{10}
\providecommand{\url}[1]{\texttt{#1}}
\providecommand{\urlprefix}{URL }
\providecommand{\doi}[1]{https://doi.org/#1}

\bibitem{afonso2010augmented}
Afonso, M.V., Bioucas-Dias, J.M., Figueiredo, M.A.: An augmented lagrangian
  approach to the constrained optimization formulation of imaging inverse
  problems. IEEE Transactions on Image Processing  \textbf{20}(3),  681--695
  (2010)

\bibitem{aghdam2018combination}
Aghdam, M.A., Sharifi, A., Pedram, M.M.: Combination of rs-fmri and smri data
  to discriminate autism spectrum disorders in young children using deep belief
  network. Journal of digital imaging  \textbf{31}(6),  895--903 (2018)

\bibitem{assaf2008diffusion}
Assaf, Y., Pasternak, O.: Diffusion tensor imaging (dti)-based white matter
  mapping in brain research: a review. Journal of molecular neuroscience
  \textbf{34}(1),  51--61 (2008)

\bibitem{banerjee2008spectrum}
Banerjee, A., Jost, J.: On the spectrum of the normalized graph laplacian.
  Linear algebra and its applications  \textbf{428}(11-12),  3015--3022 (2008)

\bibitem{behrens2007probabilistic}
Behrens, T.E., Berg, H.J., Jbabdi, S., Rushworth, M.F., Woolrich, M.W.:
  Probabilistic diffusion tractography with multiple fibre orientations: What
  can we gain? Neuroimage  \textbf{34}(1),  144--155 (2007)

\bibitem{behzadi2007component}
Behzadi, Y., et~al.: A component based noise correction method (compcor) for
  bold and perfusion based fmri. Neuroimage  \textbf{37}(1),  90--101 (2007)

\bibitem{bennett2013advances}
Bennett, I.J., Rypma, B.: Advances in functional neuroanatomy: a review of
  combined dti and fmri studies in healthy younger and older adults.
  Neuroscience \& Biobehavioral Reviews  \textbf{37}(7),  1201--1210 (2013)

\bibitem{cabral2017functional}
Cabral, J., Kringelbach, M.L., Deco, G.: Functional connectivity dynamically
  evolves on multiple time-scales over a static structural connectome: Models
  and mechanisms. NeuroImage  \textbf{160},  84--96 (2017)

\bibitem{cai2017estimation}
Cai, B., Zille, P., Stephen, J.M., Wilson, T.W., Calhoun, V.D., Wang, Y.P.:
  Estimation of dynamic sparse connectivity patterns from resting state fmri.
  IEEE transactions on medical imaging  \textbf{37}(5),  1224--1234 (2017)

\bibitem{damaraju2014dynamic}
Damaraju, E., Allen, E.A., Belger, A., Ford, J.M., McEwen, S., Mathalon, D.,
  Mueller, B., Pearlson, G., Potkin, S., Preda, A., et~al.: Dynamic functional
  connectivity analysis reveals transient states of dysconnectivity in
  schizophrenia. NeuroImage: Clinical  \textbf{5},  298--308 (2014)

\bibitem{d2020joint}
D'Souza, N., Nebel, M., Wymbs, N., Mostofsky, S., Venkataraman, A.: A joint
  network optimization framework to predict clinical severity from resting
  state functional mri data. NeuroImage  \textbf{206},  116314 (2020)

\bibitem{dziuk2007dyspraxia}
Dziuk, M., Larson, J.G., Apostu, A., Mahone, E.M., Denckla, M.B., Mostofsky,
  S.H.: Dyspraxia in autism: association with motor, social, and communicative
  deficits. Developmental Medicine \& Child Neurology  \textbf{49}(10),
  734--739 (2007)

\bibitem{d2018generative}
D’Souza, N.S., Nebel, M.B., Wymbs, N., Mostofsky, S., Venkataraman, A.: A
  generative-discriminative basis learning framework to predict clinical
  severity from resting state functional mri data. In: International Conference
  on Medical Image Computing and Computer-Assisted Intervention. pp. 163--171.
  Springer (2018)

\bibitem{d2019coupled}
D’Souza, N.S., Nebel, M.B., Wymbs, N., Mostofsky, S., Venkataraman, A.: A
  coupled manifold optimization framework to jointly model the functional
  connectomics and behavioral data spaces. In: International Conference on
  Information Processing in Medical Imaging. pp. 605--616. Springer (2019)

\bibitem{d2019integrating}
D’Souza, N.S., Nebel, M.B., Wymbs, N., Mostofsky, S., Venkataraman, A.:
  Integrating neural networks and dictionary learning for multidimensional
  clinical characterizations from functional connectomics data. In:
  International Conference on Medical Image Computing and Computer-Assisted
  Intervention. pp. 709--717. Springer (2019)

\bibitem{everson1998orthogonal}
Everson, R.: Orthogonal, but not orthonormal, procrustes problems. Advances in
  computational Mathematics  \textbf{3}(4) (1998)

\bibitem{jenkinson2012fsl}
Jenkinson, M., Beckmann, C.F., Behrens, T.E., Woolrich, M.W., Smith, S.M.: Fsl.
  Neuroimage  \textbf{62}(2),  782--790 (2012)

\bibitem{kawahara2017brainnetcnn}
Kawahara, J., et~al.: Brainnetcnn: convolutional neural networks for brain
  networks; towards predicting neurodevelopment. NeuroImg  \textbf{146},
  1038--1049 (2017)

\bibitem{kingma2015adam}
Kingma, D.P., Ba, J.L.: Adam: A method for stochastic optimization  (2015)

\bibitem{lee2013resting}
Lee, M.H., Smyser, C.D., Shimony, J.S.: Resting-state fmri: a review of methods
  and clinical applications. American Journal of neuroradiology
  \textbf{34}(10),  1866--1872 (2013)

\bibitem{manton2003geometry}
Manton, J.H., Mahony, R., Hua, Y.: The geometry of weighted low-rank
  approximations. IEEE Transactions on Signal Processing  \textbf{51}(2),
  500--514 (2003)

\bibitem{mostofsky2006developmental}
Mostofsky, S.H., Dubey, P., Jerath, V.K., Jansiewicz, E.M., Goldberg, M.C.,
  Denckla, M.B.: Developmental dyspraxia is not limited to imitation in
  children with autism spectrum disorders. Journal of the International
  Neuropsychological Society  \textbf{12}(3),  314--326 (2006)

\bibitem{nandakumar2018defining}
Nandakumar, N., D’Souza, N.S., Craley, J., Manzoor, K., Pillai, J.J., Gujar,
  S.K., Sair, H.I., Venkataraman, A.: Defining patient specific functional
  parcellations in lesional cohorts via markov random fields. In: Connectomics
  in NeuroImaging: Second International Workshop, CNI 2018, Held in Conjunction
  with MICCAI 2018, Granada, Spain, September 20, 2018, Proceedings 2. pp.
  88--98. Springer (2018)

\bibitem{nebel2016intrinsic}
Nebel, M.B., et~al.: Intrinsic visual-motor synchrony correlates with social
  deficits in autism. Bio. Psych.  \textbf{79}(8),  633--641 (2016)

\bibitem{payakachat2012autism}
Payakachat, N., et~al.: Autism spectrum disorders: a review of measures for
  clinical, health services and cost-effectiveness applications. Expert review
  of pharmacoeconomics \& outcomes research  \textbf{12}(4),  485--503 (2012)

\bibitem{pouw2013link}
Pouw, L.B., Rieffe, C., Stockmann, L., Gadow, K.D.: The link between emotion
  regulation, social functioning, and depression in boys with asd. Research in
  Autism Spectrum Disorders  \textbf{7}(4),  549--556 (2013)

\bibitem{price2014multiple}
Price, T., Wee, C.Y., Gao, W., Shen, D.: Multiple-network classification of
  childhood autism using functional connectivity dynamics. In: International
  Conference on Medical Image Computing and Computer-Assisted Intervention. pp.
  177--184. Springer (2014)

\bibitem{rabany2019dynamic}
Rabany, L., Brocke, S., Calhoun, V.D., Pittman, B., Corbera, S., Wexler, B.E.,
  Bell, M.D., Pelphrey, K., Pearlson, G.D., Assaf, M.: Dynamic functional
  connectivity in schizophrenia and autism spectrum disorder: Convergence,
  divergence and classification. NeuroImage: Clinical  \textbf{24},  101966
  (2019)

\bibitem{rashid2014dynamic}
Rashid, B., Damaraju, E., Pearlson, G.D., Calhoun, V.D.: Dynamic connectivity
  states estimated from resting fmri identify differences among schizophrenia,
  bipolar disorder, and healthy control subjects. Frontiers in human
  neuroscience  \textbf{8}, ~897 (2014)

\bibitem{schnabel1983forcing}
Schnabel, R.B., Toint, P.L.: Forcing sparsity by projecting with respect to a
  non-diagonally weighted frobenius norm. Mathematical Programming
  \textbf{25}(1),  125--129 (1983)

\bibitem{skudlarski2008measuring}
Skudlarski, P., Jagannathan, K., Calhoun, V.D., Hampson, M., Skudlarska, B.A.,
  Pearlson, G.: Measuring brain connectivity: diffusion tensor imaging
  validates resting state temporal correlations. Neuroimage  \textbf{43}(3),
  554--561 (2008)

\bibitem{sridharan2008critical}
Sridharan, D., et~al.: A critical role for the right fronto-insular cortex in
  switching between central-executive and default-mode networks. Proc. Nat.
  Acad. Sci.  \textbf{105}(34),  12569--12574 (2008)

\bibitem{sui2013combination}
Sui, J., He, H., Yu, Q., Rogers, J., Pearlson, G., Mayer, A.R., Bustillo, J.,
  Canive, J., Calhoun, V.D., et~al.: Combination of resting state fmri, dti,
  and smri data to discriminate schizophrenia by n-way mcca+ jica. Frontiers in
  human neuroscience  \textbf{7}, ~235 (2013)

\bibitem{tzourio2002automated}
Tzourio-Mazoyer, N., et~al.: Automated anatomical labeling of activations in
  spm using a macroscopic anatomical parcellation of the mni mri single-subject
  brain. Neuroimage  \textbf{15}(1),  273--289 (2002)

\end{thebibliography}

\end{document}

\title{Supplementary Results}
\author{N.S. D'Souza et al. }

\institute{Dept. of Electrical and Computer Eng., Johns Hopkins University, Baltimore, USA
}

\authorrunning{Niharika Shimona DSouza et al.}
\maketitle    

\begin{figure} [!h]
  \centerline{\includegraphics[scale=0.33]{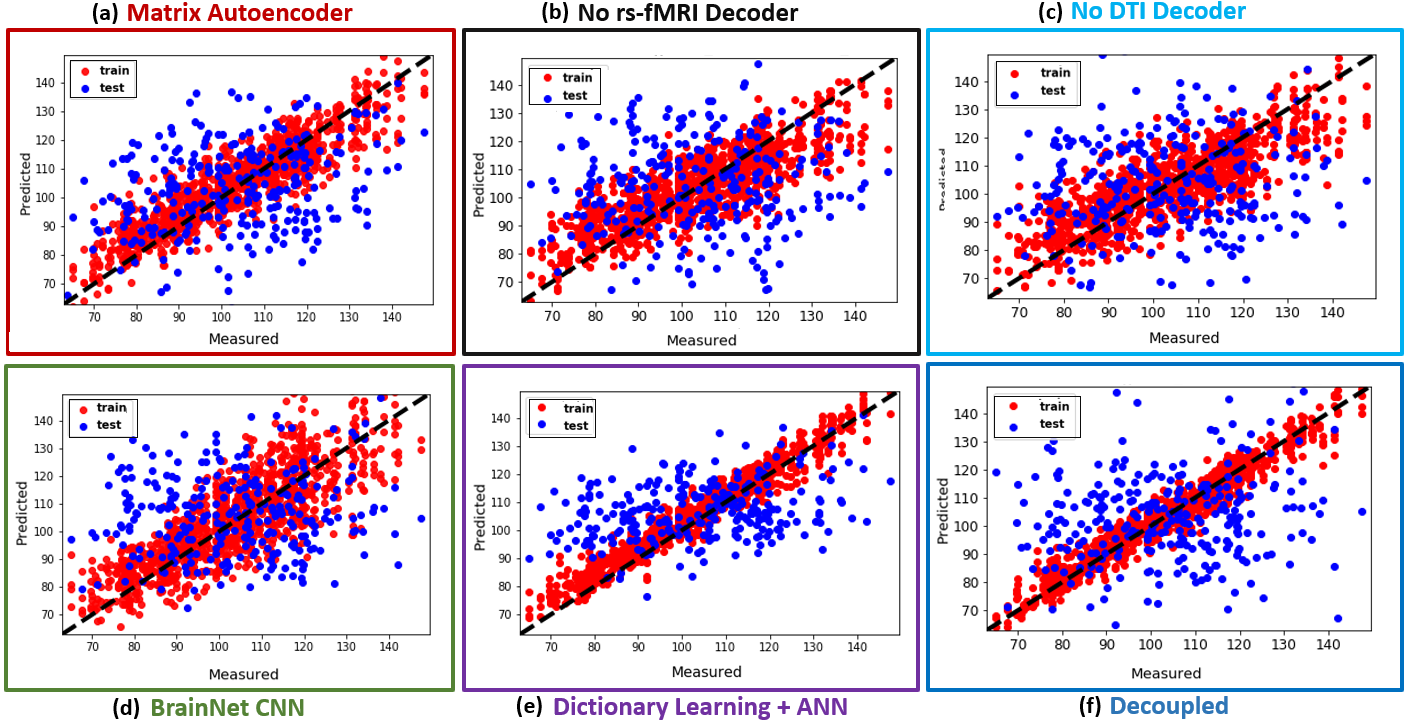}}
       \caption{{\textbf{HCP Dataset} Prediction of CFIS by \textbf{(a)} Our Framework \textbf{(b)} Matrix AE without rs-fMRI Decoder \textbf{(c)} Matrix AE without DTI Decoder \textbf{(d)} BrainNet CNN  \textbf{(e)} Dictionary Learning + ANN \textbf{(f)} Decoupled Matrix AE and ANN }}
       \label{HCP}
\end{figure}
\noindent
\begin{figure}[b!]
 \centerline{\includegraphics[scale=0.40]{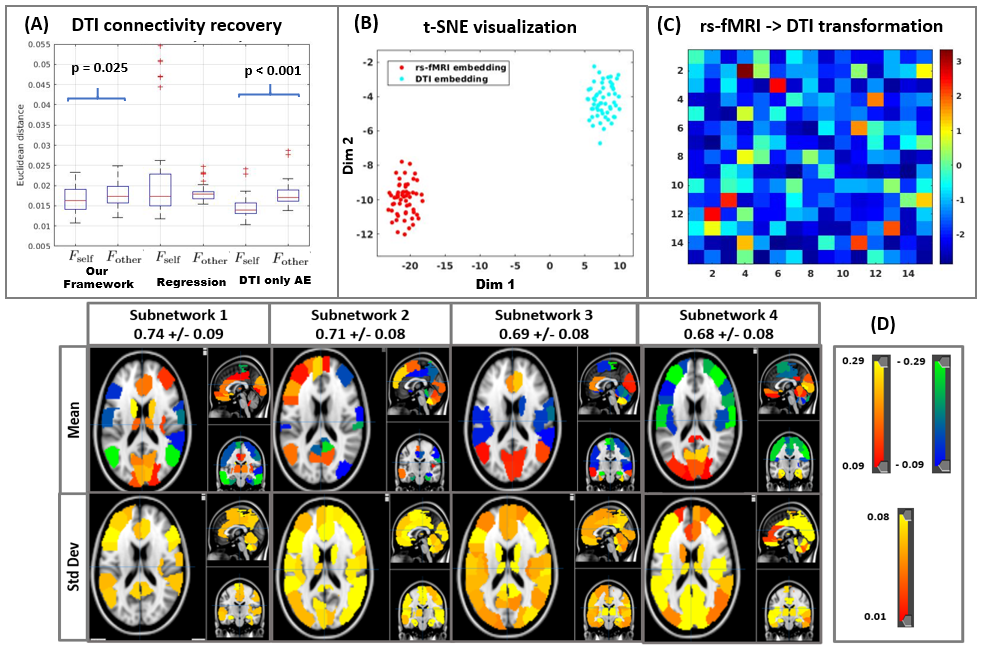}}
 \caption{{\textbf{ASD Dataset:} \textbf{(A)} SC recovery by \textbf{(L):} Our Framework \textbf{(M):} Linear Regression \textbf{(R):} DTI only AE \textbf{(B)} t-SNE visualization of embeddings \textbf{(C)} Coeff. of Var. $(C_{v})$ (log scale) for $\mathbf{\Phi}_{\text{align}}(\cdot)$. Cold colors imply better stability \textbf{(D)} Top four FC bases}}
\label{Results}
\end{figure}

\begin{figure} 
    \centering
      \includegraphics[scale = 0.35]{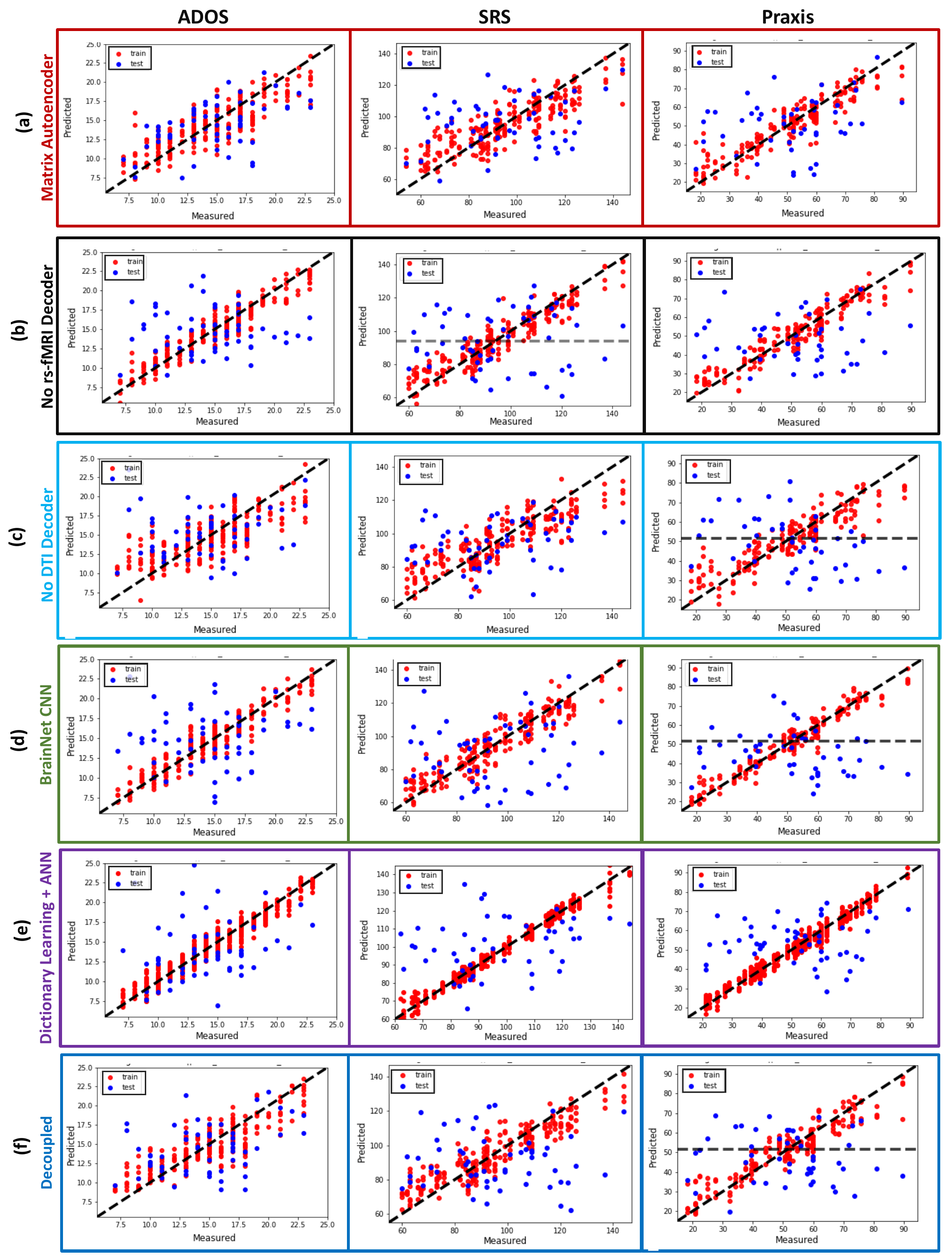}
   \caption{{\textbf{ASD Dataset:} Multi-output prediction performance of \textbf{(L):} ADOS \textbf{(M):} SRS \textbf{(R):} Praxis by \textbf{(a)} Our Framework \textbf{(b)} Matrix Autoencoder without rs-fMRI Decoder \textbf{(c)} Matrix Autoencoder without DTI Decoder \textbf{(d)} BrainNet CNN  \textbf{(e)} Dictionary Learning + ANN \textbf{(f)} Decoupled Matrix Autoencoder and ANN }}\label{KKI}
\end{figure}